%% file: main.tex
\documentclass{article}

\usepackage[preprint]{neurips_2024}

\usepackage[utf8]{inputenc}
\usepackage[T1]{fontenc}
\usepackage{amsmath, amssymb, amsfonts}
\usepackage{graphicx}
\usepackage{booktabs}
\usepackage{hyperref}
\usepackage{url}
\usepackage{nicefrac}
\usepackage{microtype}
\usepackage{xcolor}
\usepackage{cleveref}
\usepackage{enumitem}

\title{The Compression Gap: Why Discrete Tokenization Limits\\Vision-Language-Action Model Scaling}

\author{
  Takuya Shiba \\
  Shibattic Inc.\\
  \texttt{Project page: \url{https://shibattic.github.io/compression-gap}}
}

\date{}

\begin{document}
\maketitle

\begin{figure}[!h]
  \centering
  \includegraphics[width=\textwidth]{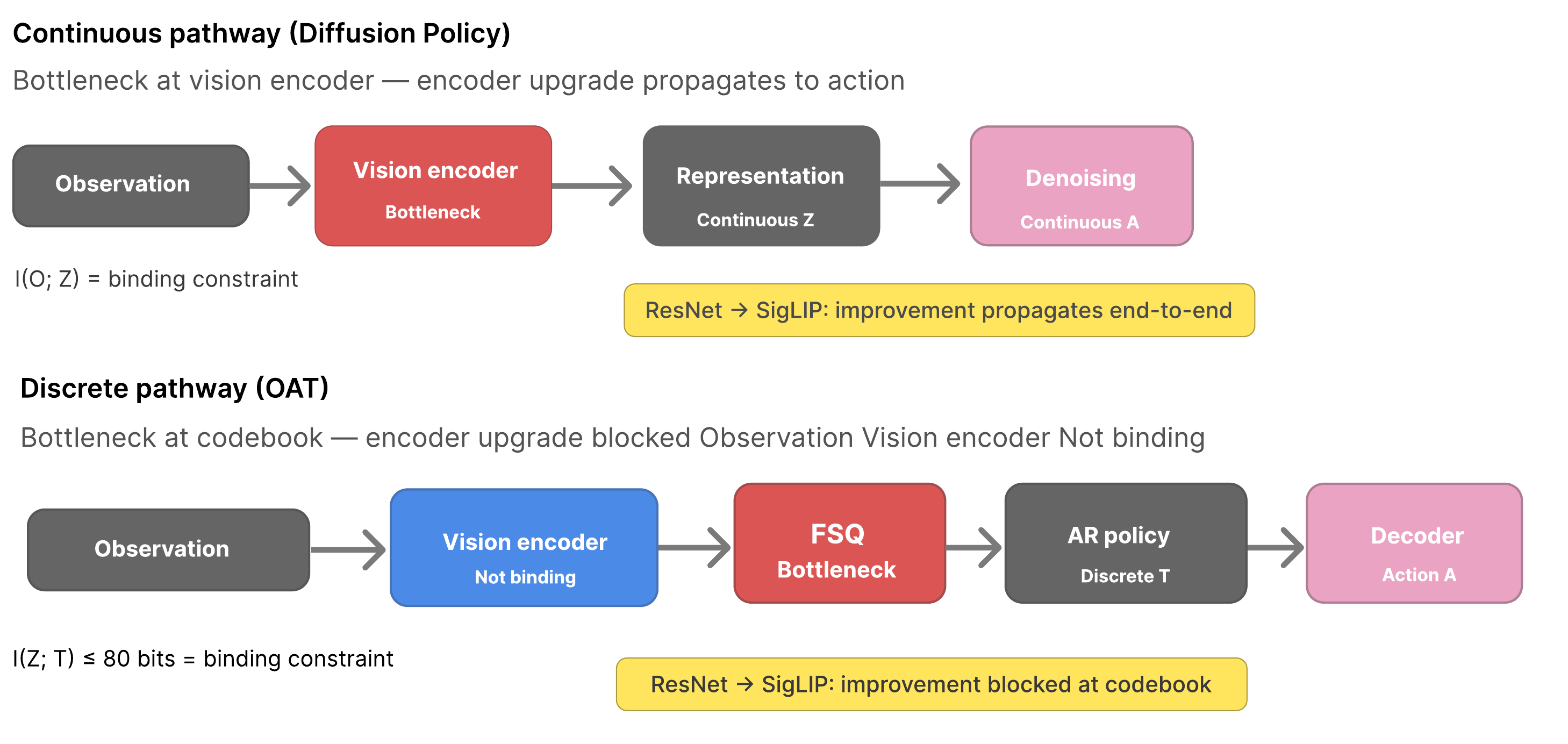}
  \caption{The Compression Gap. In the continuous pathway (Diffusion Policy), the vision encoder is the binding bottleneck; upgrading it propagates end-to-end. In the discrete pathway (OAT), the codebook ($I(Z;T) \leq 80$ bits) is the binding bottleneck; encoder upgrades are blocked at quantization.}
  \label{fig:pipeline}
\end{figure}

\begin{abstract}
Scaling Vision-Language-Action (VLA) models by upgrading the vision encoder is expected to improve downstream manipulation performance---as it does in vision-language modeling. We show that this expectation fails when actions are represented as discrete tokens, and explain why through an information-theoretic principle we call the Compression Gap: in any visuomotor pipeline, scaling behavior is governed by the location of the tightest information bottleneck. When actions are continuous (e.g., Diffusion Policy), the vision encoder is the binding constraint, and upgrading it directly improves performance. When actions are discretized through a fixed-capacity codebook (e.g., OAT), the codebook becomes the binding constraint, and encoder improvements cannot propagate past it---regardless of how rich the upstream representation is. We validate this principle on the LIBERO benchmark with three lines of evidence: a factorial experiment showing that encoder upgrades improve Diffusion Policy by over 21 percentage points while OAT gains are substantially attenuated across model scales; an encoder quality gradient across four encoders confirming that Diffusion Policy tracks encoder quality monotonically while OAT remains flat; and a codebook size experiment demonstrating that relaxing codebook capacity partially recovers encoder sensitivity, providing causal evidence for the bottleneck hypothesis. Our findings reveal that scaling in Physical AI requires identifying where information bottlenecks lie in the pipeline, rather than uniformly increasing model or data size.
\end{abstract}

\input{introduction}

\input{background}
\input{experiments}
\input{results}
\input{conclusion}

\clearpage
\bibliographystyle{plainnat}
\bibliography{references}


\end{document}

%% file: introduction.tex
\section{Introduction}
\label{sec:introduction}

A central premise of scaling in Vision-Language-Action (VLA) models is that improving the vision encoder improves downstream task performance. This assumption is well-supported in vision-language modeling: recent large-scale studies \citep{tong2026scaling} demonstrate that higher-quality visual representations consistently translate into better understanding and generation capabilities across modalities. A natural expectation is that the same principle holds for robotic manipulation---that upgrading from a task-agnostic encoder (e.g., ResNet-18) to a semantically richer one (e.g., SigLIP) should yield better policies, regardless of how actions are represented.

We show that this expectation is false in general. Whether a vision encoder upgrade improves manipulation performance depends critically on the choice of action representation. We compare two representative families---discrete tokenization (OAT) and continuous diffusion (Diffusion Policy)---under identical encoder upgrades on the LIBERO benchmark. Diffusion Policy absorbs the full benefit of the encoder upgrade, while OAT exhibits substantially weaker and less consistent sensitivity. This gap persists across model scales (M/L), ruling out confounds from capacity.

We explain this phenomenon through the lens of information bottleneck location. In continuous action representations, the tightest bottleneck lies at the vision encoder itself; upgrading it directly widens the information pathway to action prediction. In discrete representations, the bottleneck lies at the action tokenizer---a fixed-capacity codebook that truncates information regardless of encoder quality. Improvements upstream of a binding bottleneck cannot propagate past it. (Figure~\ref{fig:pipeline}) We term this phenomenon the \textbf{Compression Gap}: the divergence in scaling behavior that arises from the location of the information bottleneck within the visuomotor pipeline.

This finding connects to a broader pattern. \citet{tong2026scaling} independently observe at pretraining scale that semantic encoders (RAE) continue to improve with increased model capacity while VAE-based representations saturate---the same structural asymmetry we identify at the manipulation-policy scale. The convergence suggests that the Compression Gap is not an artifact of a particular benchmark or architecture, but may reflect a broader consequence of how lossy compression stages interact with scaling.

Specifically, we investigate the following axes:

\begin{itemize}[leftmargin=*, itemsep=3pt]
    \item \textbf{Vision Encoder Sensitivity.} Under identical encoder upgrades (ResNet-18 $\rightarrow$ SigLIP), Diffusion Policy improves substantially while OAT shows markedly attenuated sensitivity. This interaction effect is the primary evidence for the Compression Gap. (\S4.1)
    \item \textbf{Encoder Quality Gradient.} We extend beyond the binary ResNet/SigLIP comparison by evaluating multiple vision encoders of varying quality (DINOv2, SigLIP 2). Diffusion Policy's success rate tracks encoder quality monotonically; OAT's remains flat---providing continuous evidence for the bottleneck hypothesis. (\S4.2)
    \item \textbf{Codebook Size.} We vary the codebook capacity of OAT and measure how encoder sensitivity changes. If the codebook is the binding bottleneck, increasing its capacity should recover sensitivity to encoder quality---providing causal evidence for the bottleneck hypothesis. (\S4.3)
\end{itemize}

%% file: background.tex
\section{Background}
\label{sec:background-related-work}

Our goal is to identify whether improvements in vision encoder quality propagate to manipulation performance, and if so, under what conditions. To isolate the effect of action representation, we compare two families---discrete tokenization (OAT) and continuous diffusion (Diffusion Policy)---that share the same observation pipeline but differ in how actions are encoded. We first introduce each representation and formalize the information pathway from observation to action (\S2.1--\S2.2), then position our work relative to prior research (\S2.3).

\subsection{Action Representations for Visuomotor Policy}
\label{sec:background-action-representations}

We compare two representative families of action representation that share the same observation pipeline but differ fundamentally in how actions are encoded: discrete tokenization via OAT and continuous denoising via Diffusion Policy. Both predict an action chunk $a_{1:H_a}$ of horizon $H_a$ conditioned on an observation history $o_{1:H_o}$.

\textbf{Discrete: Ordered Action Tokenization (OAT).} OAT \citep{liu2026oat} is a two-stage framework. First, a learned tokenizer $\mathcal{T}$ compresses a continuous action chunk into a short sequence of discrete tokens:

\[
\mathcal{T}: a_{1:H_a} \mapsto T_{1:H_l}, \quad T_i \in \mathcal{V}.
\]

where $H_l \ll H_a$ is the latent token horizon and $\mathcal{V}$ is a finite vocabulary. Internally, a transformer encoder with learnable register tokens aggregates the action sequence into $H_l$ latent vectors, which are then discretized via Finite Scalar Quantization (FSQ; \citealp{mentzer2024finite}) with levels $[8, 5, 5, 5]$, yielding a codebook of size $|\mathcal{V}| = 1000$. A decoder $\mathcal{T}^{-1}$ maps token sequences back to continuous actions. Second, an autoregressive policy generates tokens sequentially:

\[
p(T_{1:H_l} \mid o_{1:H_o}) = \prod_{i=1}^{H_l} p(T_i \mid T_{<i}, o_{1:H_o}).
\]

The critical feature for our analysis is the FSQ bottleneck: regardless of the richness of the observation representation, all action information must pass through a discrete codebook of fixed capacity before reaching execution.

\textbf{Continuous: Diffusion Policy (DP).} Diffusion Policy \citep{chi2024diffusion} models the action distribution as a denoising process in continuous space. Starting from Gaussian noise $a^T \sim \mathcal{N}(0, I)$, the policy iteratively denoises toward a clean action chunk:

\[
a^{t-1} = \mathrm{Denoise}_{\theta}(a^t, t, o_{1:H_o}), \quad t = T, \ldots, 1.
\]

using a learned noise prediction network $\epsilon_{\theta}$ with a DDIM scheduler. The entire pipeline operates in continuous space: observations are encoded into continuous features, and actions are predicted as continuous vectors without any discretization stage.

\textbf{Key structural difference.}

Both representations map observations to actions, but discrete methods insert a fixed-capacity codebook bottleneck. In OAT, action chunks are quantized into a finite FSQ vocabulary ($|\mathcal{V}| = 1000$ by default), so additional encoder information is discarded once the codebook saturates. Diffusion Policy has no discrete quantization stage, so encoder information flows continuously to actions.

\subsection{Information Bottleneck in Visuomotor Pipelines}
\label{sec:background-information-bottleneck}

We now formalize why the choice of action representation determines whether vision encoder improvements propagate to task performance. Consider the information pathway common to all visuomotor policies:

\[
O \rightarrow Z \rightarrow A.
\]

where $O$ denotes the raw observation, $Z$ the learned representation (the output of the vision encoder and any intermediate processing), and $A$ the executed action. By the data processing inequality \citep{cover2006elements}, for any such Markov chain:

\[
I(O; A) \leq \min\left(I(O; Z),\, I(Z; A)\right).
\]

That is, the mutual information between observation and action is bounded by the tightest bottleneck along the pathway. Improving one stage can only increase end-to-end information flow if that stage is the binding constraint.

For visuomotor policies, it is useful to decompose the pathway further. Both OAT and Diffusion Policy share the same observation encoder $f_{\mathrm{enc}}$, which maps raw observations to a continuous representation $Z = f_{\mathrm{enc}}(O)$. The two representations diverge in how $Z$ is mapped to actions.

\textbf{Continuous pathway (DP).} Diffusion Policy maps $Z$ to actions through a continuous denoising process:

\[
O \xrightarrow{f_{\mathrm{enc}}} Z \xrightarrow{\epsilon_{\theta}} A.
\]

Every stage operates in continuous space. The mutual information $I(Z; A)$ is not subject to any hard combinatorial bound---it is limited only by the capacity of the denoising network $\epsilon_{\theta}$ and the quality of $Z$ itself. Consequently, the binding bottleneck is typically $I(O; Z)$: the vision encoder determines how much task-relevant information enters the pipeline. Upgrading the encoder directly widens this bottleneck, and the improvement flows through to action prediction.

\textbf{Discrete pathway (OAT and other discrete tokenizers).} Discrete representations introduce a quantization stage $Q$ between representation and action:

\[
O \xrightarrow{f_{\mathrm{enc}}} Z \xrightarrow{Q} T \xrightarrow{\mathcal{T}^{-1}} A,
\]

where $T \in \mathcal{V}^{H_l}$ is the discrete token sequence and $Q$ denotes the quantization operation (FSQ in OAT, BPE in FAST, binning in Bin). Applying the data processing inequality to this extended chain:

\[
I(O; A) \leq I(Z; T) \leq \log_2 |\mathcal{V}|^{H_l} = H_l \log_2 |\mathcal{V}|.
\]

The discrete codebook imposes a hard upper bound on information throughput. For OAT with $|\mathcal{V}| = 1000$ and $H_l = 8$, this bound is approximately 80 bits per action chunk. If this capacity is already saturated by the existing encoder, upgrading to a richer encoder cannot increase $I(O; A)$---the additional information is discarded at the quantization stage. The binding bottleneck is $I(Z; T)$, not $I(O; Z)$.

\textbf{Prediction.} This analysis yields a testable prediction. Let $\Delta_{\mathrm{enc}}$ denote the change in task performance when upgrading the vision encoder from a weaker encoder (ResNet-18) to a stronger one (SigLIP). For the continuous pathway, $\Delta_{\mathrm{enc}} > 0$ because the binding bottleneck ($I(O; Z)$) is directly widened. For the discrete pathway, $\Delta_{\mathrm{enc}} \approx 0$ if the codebook is already the binding constraint---the encoder upgrade widens a non-binding bottleneck and has no effect on end-to-end information flow. (Figure~\ref{fig:pipeline}) We term this divergence the \textbf{Compression Gap}. Sections 4 and 5 test this prediction empirically.

\subsection{Related Work}
\label{sec:background-related-work-sub}

\textbf{Action representations for visuomotor policy.} A growing body of work has explored how to represent robot actions for policy learning. Diffusion Policy \citep{chi2024diffusion} models actions as a continuous denoising process and has become a widely adopted baseline. Along the same continuous trajectory, flow matching approaches such as $\pi_0$ \citep{black2024pi0} replace the diffusion process with a continuous flow field, broadening the family of continuous action representations. On the discrete side, per-dimension binning \citep{brohan2022rt1, brohan2023rt2} offers simplicity but produces long token sequences; FAST \citep{pertsch2025fast} achieves compact tokenization via frequency-domain compression but suffers from partial decodability; VQ-BeT \citep{lee2024vqbet} and QueST \citep{mete2024quest} use learned latent quantization; and OAT \citep{liu2026oat} introduces ordered tokenization satisfying compression, decodability, and causal structure simultaneously. ACT \citep{zhao2023act} takes a distinct approach with a CVAE-based continuous representation. These works focus on comparing task performance across representations. Our work asks a complementary question: how does each representation respond to improvements in the upstream vision encoder? This axis of comparison---sensitivity to encoder scaling---has not been systematically studied.

\textbf{Scaling laws and multimodal pretraining.} Scaling laws for language models \citep{kaplan2020scaling, hoffmann2022training} established that performance improves predictably with compute, and that compute-optimal allocation between parameters and data follows power-law relationships. Recent work has extended this analysis to multimodal settings. \citet{tong2026scaling} conduct from-scratch multimodal pretraining experiments and uncover a scaling asymmetry: vision is significantly more data-hungry ($D_{\text{opt}} \propto C^{0.63}$) than language ($D_{\text{opt}} \propto C^{0.53}$). Critically, they find that semantic encoders (RAE, SigLIP 2 \citep{tschannen2025siglip2}) continue to improve with increased model capacity while VAE-based representations saturate---a pattern structurally analogous to the Compression Gap we identify at the policy level. Our contribution is to show that whether encoder quality matters depends on the action representation---a finding that connects the robotics and multimodal scaling literatures.


%% file: experiments.tex
\section{Experiment Design}
\label{sec:experiment-design}

We design a controlled experiment to test the prediction derived in \S2.2: that vision encoder upgrades improve task performance under continuous action representations ($\Delta_{\mathrm{enc}} > 0$ for DP) but not under discrete ones ($\Delta_{\mathrm{enc}} \approx 0$ for OAT). To isolate this interaction effect from potential confounds, we construct a factorial experiment with three binary variables, yielding
\[
2 \times 2 \times 2 = 8
\]
conditions. We further probe the bottleneck hypothesis through two supplementary experiments: an encoder quality gradient that extends the binary encoder comparison to multiple encoders, and a codebook size variation that tests whether relaxing the discrete bottleneck recovers encoder sensitivity. All conditions share the same codebase, training procedure, and evaluation protocol; only the specified variables differ across runs.

\subsection{Experimental Variables}
\label{sec:experiment-design-variables}

We manipulate three variables in the factorial design, each chosen to test a specific aspect of the Compression Gap hypothesis.

\textbf{Action representation (OAT / Diffusion Policy).} This is the primary variable. OAT represents the discrete pathway with an FSQ codebook ($|\mathcal{V}| = 1000$, $H_l = 8$), and Diffusion Policy represents the continuous pathway with a DDIM scheduler (100 training steps, 10 inference steps). Both use the same transformer-based policy backbone, ensuring that differences in performance arise from the action representation rather than model architecture. We use the Diffusion Policy baseline included in the OAT codebase \citep{liu2026oat} for a fair comparison. We select OAT as the representative discrete tokenizer because it satisfies compression, decodability, and causal ordering simultaneously \citep{liu2026oat}, making it the strongest available discrete baseline. If the Compression Gap manifests even under the best-performing discrete tokenizer, the finding is more likely to reflect a structural property of discrete representations rather than a limitation of a specific method.

\textbf{Vision encoder (ResNet-18 / SigLIP).} This is the intervention variable. ResNet-18 \citep{he2016resnet} produces a 64-dimensional feature with spatial pooling, representing a compact but informationally limited observation encoding. SigLIP \citep{zhai2023siglip} provides a 1152-dimensional semantic representation pretrained on web-scale image-text pairs, representing a substantially richer encoding. The Compression Gap predicts that this upgrade improves performance for DP but not for OAT. For SigLIP, we use pre-cached features to ensure training efficiency.

To extend beyond this binary comparison, we conduct an additional encoder quality gradient experiment. Fixing the model size to M and the benchmark to LIBERO-10, we evaluate both action representations under four vision encoders of varying quality: ResNet-18 (64-d), SigLIP (1152-d), DINOv2 ViT-L/14 (1024-d), and SigLIP 2 (1152-d; \citealp{tschannen2025siglip2}). This experiment tests whether the Compression Gap holds continuously across encoder quality, rather than only at two points.

\textbf{Model size (M / L).} This variable controls for the possibility that the absence of encoder sensitivity in OAT is merely a capacity artifact. If OAT fails to benefit from SigLIP simply because the policy backbone is too small, then scaling to a larger model should recover the benefit. If the Compression Gap hypothesis is correct---that the codebook, not the policy capacity, is the binding bottleneck---then scaling model size should not resolve the insensitivity. Size M uses a 4-layer transformer decoder with embedding dimension 256 and 4 attention heads. Size L scales to 6 layers with embedding dimension 384 and 6 attention heads. Both configurations use a head dimension of 64.

\textbf{Codebook size variation.} To provide causal evidence for the bottleneck hypothesis, we conduct a supplementary experiment varying the codebook capacity of OAT. Fixing the model size to M and the benchmark to LIBERO-10, we train OAT under three FSQ configurations: $|\mathcal{V}| = 1000$ (levels [8,5,5,5], default), $|\mathcal{V}| = 1920$ (levels [8,8,6,5]), and $|\mathcal{V}| = 4375$ (levels [7,5,5,5,5]). For each codebook size, we train with both ResNet-18 and SigLIP and measure
\[
\Delta_{\mathrm{enc}} = SR(\mathrm{SigLIP}) - SR(\mathrm{ResNet}).
\]
If the codebook acts as a compression stage that masks encoder quality, relaxing it should cause OAT's performance under each encoder to converge toward the corresponding Diffusion Policy performance---the behavior expected when no discrete bottleneck is present.

The full experimental matrix is summarized in Table~\ref{tab:factorial}.

\subsection{Benchmark}
\label{sec:experiment-design-benchmark}

We evaluate on LIBERO \citep{liu2023libero}, built on the Franka Emika Panda, with a 7-dimensional action space (3D position, 3D orientation, 1D gripper); all policies predict $H_a = 32$ action chunks and execute the first 16 steps before re-inference. We use LIBERO-10 (10 tasks, 50 demonstrations per task), the primary benchmark in the OAT paper, for direct comparability.

\subsection{Implementation}
\label{sec:experiment-design-implementation}

All experiments use the official OAT codebase \citep{liu2026oat}. Both OAT and Diffusion Policy share the same policy backbone architecture described in \S3.1. The observation encoder pipeline is identical across all conditions. For ResNet-18, we use the default configuration from the OAT codebase with spatial softmax pooling and a 64-dimensional output. For SigLIP, we pre-extract and cache features offline to ensure training efficiency.

For the encoder quality gradient experiment, we additionally evaluate DINOv2 ViT-L/14, which produces a 1024-dimensional representation pretrained with self-supervised learning on image data \citep{oquab2024dinov2}, and SigLIP 2, which provides a 1152-dimensional representation from an updated vision-language pretraining procedure \citep{tschannen2025siglip2}. Both use the same pre-caching protocol as SigLIP.

For the codebook size experiment, we modify only the FSQ level configuration in the OAT tokenizer, retraining the tokenizer and policy from scratch for each $|\mathcal{V}|$. All other hyperparameters remain identical to the default setting.

Training uses AdamW with a constant learning rate of $5 \times 10^{-5}$ for the policy network and $1 \times 10^{-5}$ for the observation encoder, with no weight decay. All models are trained for 300 epochs on a single NVIDIA A100 GPU per run. We evaluate success rate every 50 epochs via rollouts, with 500 rollouts per evaluation (50 per task). Final performance is reported as the peak success rate across training. We adopt peak success rate to match prior practice in the OAT codebase and to compare best-achieved policy quality across conditions; variance-aware evaluation across multiple seeds is left to future work.

%% file: results.tex
\section{Results and Analysis}
\label{sec:results-analysis}

We now test the predictions from \S2.2. Recall that the Compression Gap hypothesis predicts two divergent behaviors: for continuous representations (DP), upgrading the vision encoder should improve task performance ($\Delta_{\mathrm{enc}} > 0$) because the encoder is the binding bottleneck; for discrete representations (OAT), the same upgrade should have no effect ($\Delta_{\mathrm{enc}} \approx 0$) because the codebook is the binding bottleneck. We examine this prediction along three axes: encoder sensitivity (\S4.1), encoder quality gradient (\S4.2), and codebook size (\S4.3). All results from the factorial experiment are summarized in Table~\ref{tab:factorial}.

\subsection{Vision Encoder Sensitivity}
\label{sec:results-encoder-sensitivity}

\begin{table}[t]
\centering
\caption{Factorial experiment results on LIBERO-10. Peak success rate (\%) across all 8 conditions. $\Delta_{\mathrm{enc}}$ denotes the change in success rate when upgrading from ResNet-18 to SigLIP. The Compression Gap is visible as the stark difference in $\Delta_{\mathrm{enc}}$ between DP and OAT.}
\label{tab:factorial}
\smallskip
\begin{tabular}{llccc}
\toprule
Action Repr. & Size & ResNet-18 & SigLIP & $\Delta_{\mathrm{enc}}$ \\
\midrule
DP & M & 36.4 & 57.6 & +21.2 \\
DP & L & 44.0 & 70.0 & +26.0 \\
\midrule
OAT & M & 53.8 & 57.4 & +3.6 \\
OAT & L & 48.0 & 58.4 & +10.4 \\
\bottomrule
\end{tabular}
\end{table}

Table~\ref{tab:factorial} reports success rates across all 8 conditions. The central finding is a stark interaction effect between action representation and encoder quality.

For Diffusion Policy, upgrading from ResNet-18 to SigLIP yields substantial improvements across both model sizes. At size M, success rate increases from 36.4\% to 57.6\% ($\Delta_{\mathrm{enc}} = +21.2\%$). At size L, the same upgrade improves performance from 44.0\% to 70.0\% ($\Delta_{\mathrm{enc}} = +26.0\%$). The encoder upgrade consistently benefits the continuous pathway regardless of model scale.

For OAT, the same encoder upgrade produces substantially attenuated gains. At size M, success rate moves from 53.8\% to 57.4\% ($\Delta_{\mathrm{enc}} = +3.6\%$). At size L, performance moves from 48.0\% to 58.4\% ($\Delta_{\mathrm{enc}} = +10.4\%$). Despite receiving a substantially richer observation representation, the discrete pathway shows markedly smaller gains than the continuous pathway.

This result directly confirms the prediction of \S2.2. In the continuous pathway, the vision encoder is the binding bottleneck: widening it via SigLIP increases $I(O; Z)$ and the improvement propagates to task success. In the discrete pathway, the codebook constrains information flow: the encoder upgrade increases $I(O; Z)$ but
\[
I(Z; T) \leq 80
\]
bits remains the tighter constraint, attenuating the improvement that reaches execution.

Model size scaling further supports this interpretation. For Diffusion Policy, M$\rightarrow$L yields a large gain with SigLIP (57.6\% $\rightarrow$ 70.0\%) but only a modest gain with ResNet (36.4\% $\rightarrow$ 44.0\%)---when the encoder is the binding constraint, scaling the decoder has limited effect, consistent with \S2.2. OAT shows no consistent scaling gain across encoders (Table~\ref{tab:factorial}), indicating that added autoregressive capacity cannot bypass the codebook bottleneck.

\subsection{Encoder Quality Gradient}
\label{sec:results-encoder-quality-gradient}

\begin{table}[t]
\centering
\caption{Encoder quality gradient on LIBERO-10 (M size). DP's success rate tracks encoder quality monotonically; OAT's remains in a narrow band regardless of encoder.}
\label{tab:encoder-gradient}
\smallskip
\begin{tabular}{lccc}
\toprule
Encoder & Dim & DP & OAT \\
\midrule
ResNet-18 & 64 & 36.4 & 53.8 \\
SigLIP & 1152 & 57.6 & 57.4 \\
SigLIP 2 & 1152 & 62.8 & 44.2 \\
DINOv2 ViT-L/14 & 1024 & 63.8 & 51.0 \\
\bottomrule
\end{tabular}
\end{table}

The factorial design in \S4.1 establishes the Compression Gap at two encoder quality points. To test whether this divergence holds continuously, we evaluate both action representations under four vision encoders of varying quality, fixing the model size to M and the benchmark to LIBERO-10.

Table~\ref{tab:encoder-gradient} reports the results. Diffusion Policy's success rate tracks encoder quality monotonically: 36.4\% (ResNet-18), 57.6\% (SigLIP), 62.8\% (SigLIP 2), and 63.8\% (DINOv2 ViT-L/14). Each encoder upgrade translates directly into improved task performance, consistent with the encoder being the binding bottleneck in the continuous pathway.

OAT's success rate shows no systematic relationship with encoder quality: 53.8\% (ResNet-18), 57.4\% (SigLIP), 44.2\% (SigLIP 2), and 51.0\% (DINOv2 ViT-L/14). Despite an order-of-magnitude increase in representational dimensionality from ResNet-18 to DINOv2, OAT's performance fluctuates without a clear trend. The codebook bottleneck prevents upstream improvements from propagating, regardless of how large the encoder improvement is.

The divergence between the two pathways widens with encoder quality. At the weakest encoder (ResNet-18), OAT outperforms DP by 17.4 percentage points. At the strongest encoder (DINOv2), DP surpasses OAT by 12.8 percentage points---a reversal of nearly 30 points. This crossover demonstrates that the relative advantage of discrete versus continuous representations is not fixed, but depends on the quality of the upstream encoder. When encoder quality is low, OAT's structured tokenization compensates for limited perceptual information. When encoder quality is high, only the continuous pathway can exploit the additional information.

We also note that Diffusion Policy's performance plateaus in the low 60s with the strongest encoders. This suggests that once the discrete bottleneck is removed, the encoder itself becomes the next binding constraint---and that existing vision encoders, trained for semantic understanding or self-supervised representation learning, may not capture all information relevant to manipulation. While the exact OAT values are not monotonic across all encoder settings, the central pattern is robust: Diffusion Policy tracks encoder quality systematically, while OAT does not. We return to this observation in \S5.

\subsection{Codebook Size}
\label{sec:results-codebook-size}

\begin{table}[t]
\centering
\caption{Codebook size experiment on LIBERO-10 (M size). As codebook capacity increases, OAT's behavior under each encoder converges toward DP's corresponding performance, providing causal evidence that the codebook mediates the relationship between encoder quality and task performance. DP reference: ResNet 36.4\%, SigLIP 57.6\%.}
\label{tab:codebook}
\smallskip
\begin{tabular}{lcccccc}
\toprule
$|\mathcal{V}|$ & FSQ Levels & Bits/chunk & ResNet-18 & SigLIP & $\Delta_{\mathrm{enc}}$ \\
\midrule
1000 & [8,5,5,5] & $\sim$80 & 53.8 & 57.4 & +3.6 \\
1920 & [8,8,6,5] & $\sim$87 & 42.6 & 57.8 & +15.2 \\
4375 & [7,5,5,5,5] & $\sim$97 & 54.6 & 58.6 & +4.0 \\
\bottomrule
\end{tabular}
\end{table}

The results in \S4.1 and \S4.2 establish that OAT is insensitive to encoder quality, consistent with the codebook being the binding bottleneck. However, this evidence is correlational. To provide causal evidence, we test whether relaxing the codebook bottleneck changes how OAT responds to encoder quality.

We vary the codebook capacity of OAT across three FSQ configurations---$|\mathcal{V}| = 1000$, 1920, and 4375---while training with both ResNet-18 and SigLIP under each configuration. All other variables are fixed to M size and LIBERO-10. If the codebook acts as a compression stage that masks encoder quality, relaxing it should cause OAT's performance under each encoder to converge toward the corresponding DP performance---the behavior expected when no discrete bottleneck is present.

Table~\ref{tab:codebook} reports the results alongside the DP reference values (ResNet 36.4\%, SigLIP 57.6\%). At the default $|\mathcal{V}| = 1000$, both encoders produce similar performance (ResNet 53.8\%, SigLIP 57.4\%), consistent with the codebook masking the difference in encoder quality. At $|\mathcal{V}| = 1920$, a striking asymmetry emerges: SigLIP remains stable at 57.8\%, while ResNet drops sharply to 42.6\%---approaching DP's ResNet performance of 36.4\%. The codebook compression has been partially relaxed, and the true quality of each encoder begins to be exposed: SigLIP's rich representations sustain performance, while ResNet's limited information capacity is revealed.

At $|\mathcal{V}| = 4375$, ResNet recovers to 54.6\% and SigLIP rises to 58.6\% (consistent with the modelability--capacity tradeoff in \citet{liu2026oat}). Despite this non-monotonicity, the asymmetric response at $|\mathcal{V}| = 1920$ remains the key causal evidence that codebook capacity mediates encoder effects.

%% file: conclusion.tex
\section{Discussion and Conclusion}
\label{sec:discussion-conclusion}

We set out to test whether vision encoder improvements propagate to manipulation performance, and found that the answer depends entirely on the action representation. By systematically varying encoder quality, action representation, and model size within a single controlled framework, we identify the Compression Gap: a structural divergence in scaling behavior caused by the location of the information bottleneck in the visuomotor pipeline.

\textbf{Discrete bottlenecks block upstream improvements.} When the action pathway is dominated by a fixed-capacity discrete bottleneck, encoder improvements are strongly attenuated. Replacing ResNet-18 with SigLIP---an upgrade that provides an order-of-magnitude increase in representational dimensionality---produces substantially smaller gains under OAT than under Diffusion Policy, and this attenuation persists across model scales and across a gradient of four encoder qualities (\S4.2). This is not a failure of OAT as a tokenizer; it is a structural consequence of any fixed-capacity discrete bottleneck. The information that flows from observation to action is bounded by the codebook capacity
\[
H_l \log_2 |\mathcal{V}| \approx 80
\]
bits, and no upstream improvement can substantially exceed this bound. We hypothesize that similar behavior may arise in other discrete tokenization schemes (FAST, Binning, VQ-BeT) that impose fixed-capacity bottlenecks.

\textbf{Continuous representations enable component scaling.} Diffusion Policy absorbs the full benefit of the encoder upgrade, and further benefits from model size scaling when paired with a strong encoder. This conditional scaling---where downstream improvements depend on upstream capacity---is exactly what the data processing inequality predicts when the binding bottleneck lies at the encoder rather than at the action decoder. The continuous information pathway allows each component to be improved independently, with gains propagating end-to-end. This property is essential for any system that aims to benefit from the rapid pace of improvement in vision foundation models.

\textbf{Connection to multimodal scaling asymmetry.} Our manipulation-scale results mirror \citet{tong2026scaling}: VAE latents saturate while semantic encoders (RAE/SigLIP 2) continue to scale. The analogy is direct---VAE saturation corresponds to our discrete codebook bottleneck, and RAE scaling to our continuous pathway. Seeing the same pattern in trillion-token pretraining and robot learning suggests a general structural property of information-processing pipelines.

\textbf{Discrete representations are not without merit.} Discrete actions remain valuable for unified language-action modeling with pretrained LLMs and prefix-based decoding \citep{liu2026oat}. Our codebook study (\S4.3) shows the bottleneck is codebook capacity, not discreteness itself, since larger codebooks partially recover encoder sensitivity. Future work should pursue adaptive codebooks or hybrid architectures that pair discrete reasoning with continuous action decoding.

\textbf{Limitations.} Our experiments are conducted on a single benchmark (LIBERO-10) with a focused set of vision encoders and two model sizes. While the consistency of results across model scales and the encoder quality gradient suggests robustness, validation on additional benchmarks, encoder families, and real-world settings would strengthen the conclusions.

\textbf{What's next?} The Compression Gap reveals that scaling in Physical AI is not simply a matter of increasing model or data size---it requires understanding where information bottlenecks lie in the pipeline. Our results show that once the discrete bottleneck is removed, encoder quality becomes the binding constraint (\S4.1). The encoder quality gradient experiment (\S4.2) further reveals that Diffusion Policy's performance plateaus in the low 60s even with the strongest available encoders---all of which were designed for semantic understanding (SigLIP, SigLIP 2) or self-supervised visual representation learning (DINOv2), not for robotic manipulation. This suggests that a complementary research direction is the development of vision encoders specifically optimized for the information demands of physical interaction: contact geometry, object affordances, and spatial dynamics that existing encoders may not fully capture. More broadly, our information-theoretic framework applies beyond the encoder-action interface studied here. Any lossy compression stage in a visuomotor pipeline---whether in observation encoding, action representation, or communication between modules---can create a bottleneck that blocks scaling. Identifying and removing these bottlenecks, rather than uniformly increasing capacity, may prove to be the more effective path toward scalable Physical AI.